\documentclass[10pt,twocolumn,letterpaper]{article}

\usepackage[pagenumbers]{cvpr} 

\usepackage{graphicx}
\usepackage{amsmath}
\usepackage{amssymb}
\usepackage{booktabs}

\usepackage[pagebackref,breaklinks,colorlinks]{hyperref}

\usepackage[capitalize]{cleveref}
\crefname{section}{Sec.}{Secs.}
\Crefname{section}{Section}{Sections}
\Crefname{table}{Table}{Tables}
\crefname{table}{Tab.}{Tabs.}

\def\cvprPaperID{7052} 
\def\confName{CVPR}
\def\confYear{2023}

\usepackage{booktabs}
\usepackage{multirow}
\usepackage{color, colortbl}
\usepackage[table, dvipsnames]{xcolor}
\definecolor{LightCyan}{rgb}{0.88,1,1}
\usepackage{algorithm}
\usepackage{listings}

\usepackage{etoolbox}
\makeatletter
\AfterEndEnvironment{algorithm}{\let\@algcomment\relax}
\AtEndEnvironment{algorithm}{\kern2pt\hrule\relax\vskip3pt\@algcomment}
\let\@algcomment\relax
\newcommand\algcomment[1]{\def\@algcomment{\footnotesize#1}}
\renewcommand\fs@ruled{\def\@fs@cfont{\bfseries}\let\@fs@capt\floatc@ruled
  \def\@fs@pre{\hrule height.8pt depth0pt \kern2pt}%
  \def\@fs@post{}%
  \def\@fs@mid{\kern2pt\hrule\kern2pt}%
  \let\@fs@iftopcapt\iftrue}
\makeatother
\newcommand\blfootnote[1]{\begingroup\renewcommand\thefootnote{}\footnote{#1}\addtocounter{footnote}{-1}\endgroup}

\begin{document}


\title{BiFormer: Vision Transformer with Bi-Level Routing Attention}

\author{
Lei Zhu\textsuperscript{1} \quad 
Xinjiang Wang\textsuperscript{2} \quad
Zhanghan Ke\textsuperscript{1} \quad
Wayne Zhang\textsuperscript{2} \quad 
Rynson Lau$^{1\dagger}$ \quad \\
 \textsuperscript{1} City University of Hong Kong
 \qquad \textsuperscript{2} SenseTime Research \\
 {\tt\small \{lzhu68-c,zhanghake2-c\}@my.cityu.edu.hk, \{wangxinjiang,wayne.zhang\}@sensetime.com} \\
 {\tt\small {Rynson.Lau@cityu.edu.hk}}
}

\maketitle

\blfootnote{$^\dagger$ Corresponding author.}

\begin{abstract}
As the core building block of vision transformers, attention is a powerful tool to capture long-range dependency.
However, such power comes at a cost: it incurs a huge computation burden and heavy memory footprint as pairwise token interaction across all spatial locations is computed.
A series of works attempt to alleviate this problem by introducing handcrafted and content-agnostic sparsity into attention, such as restricting the attention operation to be inside local windows, axial stripes, or dilated windows.
In contrast to these approaches, we propose a novel dynamic sparse attention via bi-level routing to enable a more flexible allocation of computations with content awareness.
Specifically, for a query, irrelevant key-value pairs are first filtered out at a coarse region level, and then fine-grained token-to-token attention is applied in the union of remaining candidate regions (\ie, routed regions).
We provide a simple yet effective implementation of the proposed bi-level routing attention, which utilizes the sparsity to save both computation and memory while involving only GPU-friendly dense matrix multiplications.
Built with the proposed bi-level routing attention, a new general vision transformer, named BiFormer, is then presented.
As BiFormer attends to a small subset of relevant tokens in a \textbf{query adaptive} manner without distraction from other irrelevant ones, it enjoys both good performance and high computational efficiency, especially in dense prediction tasks.
Empirical results across several computer vision tasks such as image classification, object detection, and semantic segmentation verify the effectiveness of our design.
Code is available at \url{https://github.com/rayleizhu/BiFormer}.
\end{abstract}
\vspace{-3mm}

\section{Introduction}

\begin{figure*}[!t]
    \centering
    \includegraphics[width=.8\linewidth]{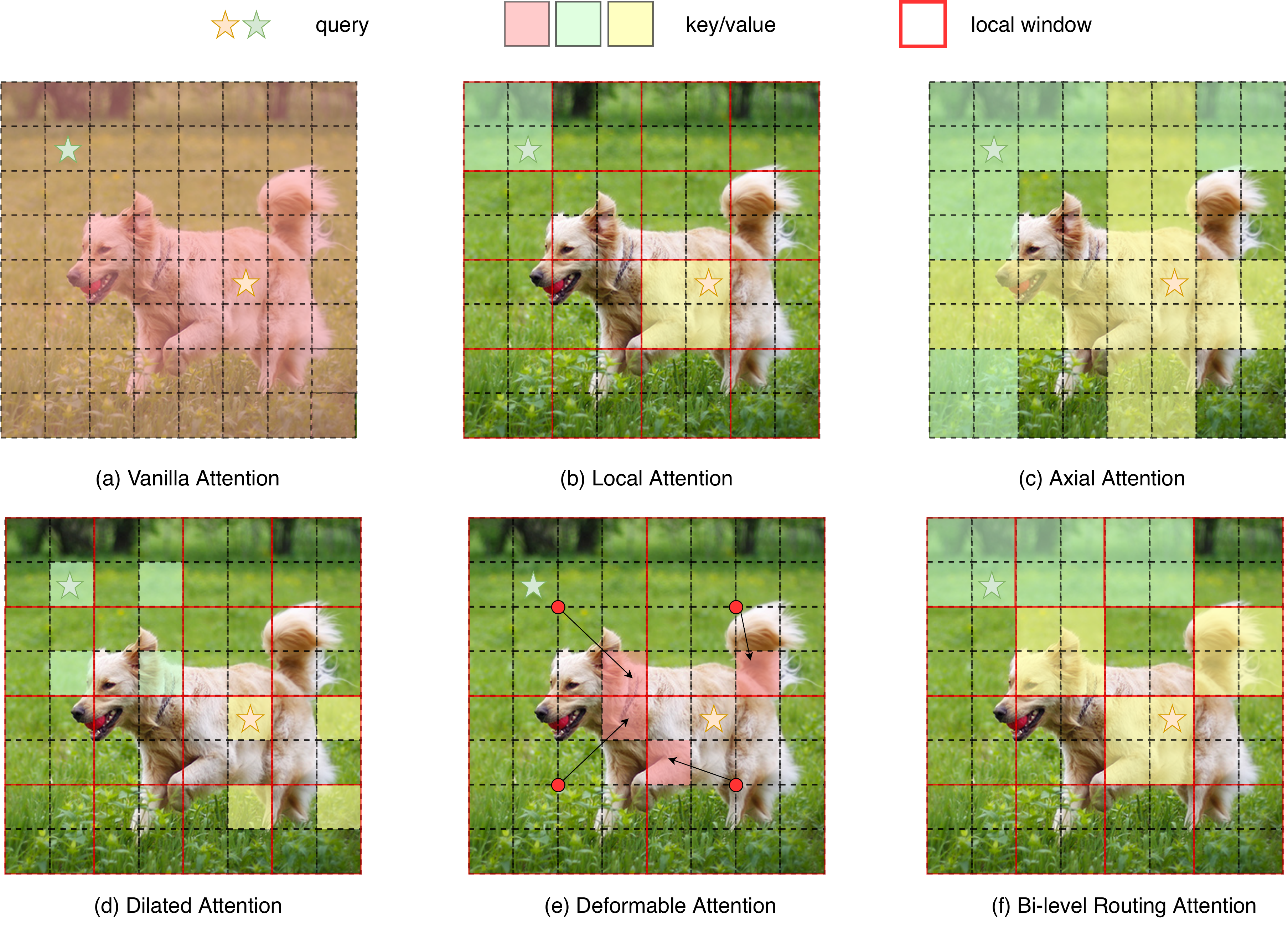}
    \vspace{-1em}
    \caption{Vanilla attention and its sparse variants. \textbf{(a)} Vanilla attention operates gloabally and incurs high computational complexity and heavy memory footprint. \textbf{(b)-(d)} Several works attempt to alleviate the complexity by introducing sparse attention with different \emph{handcrafted} patterns, such as local window~\cite{swin, wang2021crossformer}, axial stripe~\cite{cswin}, dilated window~\cite{wang2021crossformer,tu2022maxvit}. \textbf{(e)} Deformable attention~\cite{xia2022dat} enables image-adaptive sparsity via deforming a regular grid. \textbf{(f)} We achieve dynamic, query-aware sparsity with bi-level routing attention, which first searches top-$k$ ($k=3$ in this case) relevant regions, and then attends to the union of them.}
    \label{fig:attn_comparison}
    \vspace{-1em}
\end{figure*}

Transformer has many properties that are suitable for building powerful data-driven models. First, it is able to capture long-range dependency in the data~\cite{swin, vaswani2017attention}. Second, it is almost inductive-bias-free and thus makes the model more flexible to fit tons of data~\cite{vit}. Last but not least, it enjoys high parallelism, which benefits training and inference of large models~\cite{vaswani2017attention, devlin2018bert, radford2018gpt, reddy2021dalle}.
Hence, transformer has not only revolutionized natural language processing but also shown very promising progress in computer vision. 

The computer vision community has witnessed an explosion of vision transformers in the past two years~\cite{swin, cswin, wang2021pvt, vit, detr, wang2021crossformer}. Among these works, a popular topic is to improve the core building block, \ie, attention.
In contrast to convolution, which is intrinsically a local operator, a crucial property of attention is the global receptive field, which empowers vision transformers to capture long-range dependency~\cite{vaswani2017attention}.
However, such a property comes at a cost: as attention computes pairwise token affinity across all spatial locations, it has a high computational complexity and incurs heavy memory footprints.

To alleviate the problem, a promising direction is to introduce sparse attention~\cite{child2019generating} to vision transformers, so that each query attends to a small portion of key-value pairs instead of all.
In this fashion, several \emph{handcrafted} sparse patterns have been explored, such as restricting attention in local windows~\cite{swin}, dilated windows~\cite{tu2022maxvit, wang2021crossformer}, or axial stripes~\cite{wang2021crossformer}.
On the other hand, there are also works trying to make the sparsity adaptive to data~\cite{xia2022dat,chen2021dpt}. However, while they use different strategies to merge or select key/value tokens, these tokens are query-agnostic, \ie, they are shared by all queries.
Nonetheless, according to the visualization of pretrained ViT~\footnote{\url{https://epfml.github.io/attention-cnn/}}~\cite{vit} and DETR~\footnote{\url{https://colab.research.google.com/github/facebookresearch/detr/blob/colab/notebooks/detr_attention.ipynb}}~\cite{detr}, queries in different semantic regions actually attend to quite different key-value pairs. Hence, forcing all queries to attend to the same set of tokens may be suboptimal.

In this paper, we seek an attention mechanism with dynamic, query-aware sparsity.
Basically, we aim for each query to attend to a small portion of the \emph{most semantically relevant} key-value pairs.
The first problem comes as how to locate these key-value pairs to attend.
For example, if we select key-value pairs in a per-query manner as done in~\cite{gupta2021topk}, it still requires evaluation of pairwise affinity between all queries and keys, and hence has the same complexity of vanilla attention.
Another possibility is to predict attention offsets based on \emph{local context} for each query~\cite{dai2017deformable_conv,xia2022dat}, and hence pairwise affinity computation is avoided.
However, in this way, it is problematic to model long-range dependency~\cite{xia2022dat}.

To locate valuable key-value pairs to attend globally with high efficiency, we propose a region-to-region routing approach.
Our core idea is to filter out the most irrelevant key-value pairs at a coarse-grained region level, instead of directly at the fine-grained token level.
This is done by first constructing a region-level affinity graph and then pruning it to keep only top-$k$ connections for each node. Hence, each region only needs to attend to the top-$k$ routed regions.
With the attending regions determined, the next step is to apply token-to-token attention, which is non-trivial as key-value pairs are now assumed to be spatially scattered.
For this case, while the sparse matrix multiplication is applicable, it is inefficient in modern GPUs, which rely on coalesced memory operations, \ie, accessing blocks of dozens of contiguous bytes at once~\cite{nvidia_blog}.
Instead, we propose a simple solution via gathering key/value tokens, where only hardware-friendly dense matrix multiplications are involved.
We refer to this approach as \textbf{B}i-level \textbf{R}outing \textbf{A}ttention (BRA), as it contains a region-level routing step and a token-level attention step.

By using BRA as the core building block, we propose BiFormer, a general vision transformer backbone that can be used for many applications such as classification, object detection, and semantic segmentation.
As BRA enables BiFormer to attend to a small subset of the most relevant key/value tokens for each query in a content-aware manner, our model achieves a better computation-performance trade-off. 
For example, with 4.6G FLOPs computation, BiFormer-T achieves 83.8\% top-1 accuracy on ImageNet-1K classification, which is the best as far as we know under similar computation budgets without training with external data or distillation~\cite{jiang2021token_labeling, touvron2021deit}.
The improvements are also consistently shown in downstream tasks such as instance segmentation and semantic segmentation. 

To summarize, our contributions are as follows. We introduce a novel bi-level routing mechanism to vanilla attention, which enables content-aware sparse patterns in a query-adaptive manner. Using the bi-level routing attention as the basic building block, we propose a general vision transformer named BiFormer. Experimental results on various computer vision tasks including image classification, object detection, and semantic segmentation show that the proposed BiFormer achieves significantly better performances over the baselines under similar model sizes.

\section{Related Works}

\noindent
\textbf{Vision transformers.} Transformers are a family of neural networks that adopt channel-wise MLP blocks for per-location embedding (channel mixing) and attention~\cite{vaswani2017attention} blocks for cross-location relation modeling (spatial mixing). Transformers were originally proposed for natural language processing~\cite{vaswani2017attention,devlin2018bert} and then introduced to computer vision by pioneering works such as DETR~\cite{detr} and ViT~\cite{vit}. In comparison with CNNs, the biggest difference is that transformers use attention as an alternative to convolution to enable global context modeling. However, as vanilla attention computes pairwise feature affinity across all spatial locations, it incurs a high computation burden and heavy memory footprints, especially for high-resolution inputs. Hence, an important research direction is to seek more efficient attention mechanisms.

\bigbreak
\noindent
\textbf{Efficient attention mechanisms.} 
A large volume of works have been proposed to reduce the computation and memory complexity bottlenecks of vanilla attention by utilizing sparse connection patterns~\cite{child2019generating}, low-rank approximations~\cite{wang2020linformer} or recurrent operations~\cite{dai2019transformerxl}.
A thorough survey of these attention variants can be found at~\cite{tay2020transformer_survey}.
In the scope of vision transformers, sparse attention gains its popularity recently due to the tremendous success of Swin transformer~\cite{swin}.
In Swin transformer, attention is restricted to non-overlapping local windows, and the shift window operation is introduced to enable inter-window communication between \emph{adjacent} windows.
To enable larger and even quasi-global receptive fields under a reasonable computation budget, several follow-up works introduce different handcrafted sparse patterns, such as  dilated windows~\cite{wang2021crossformer,tu2022maxvit} or cross-shaped windows~\cite{cswin}.
There are also works that try to make the sparse pattern adaptive to data, such as DAT~\cite{xia2022dat}, TCFormer~\cite{zeng2022tcformer} and DPT~\cite{chen2021dpt}.
While these works reduce the number of key/value tokens via different merging or selection strategies, these key/value tokens are shared by all queries on an image.
Instead, we explore query-aware key/value token selection.
The key observation which motivates our work is that the attentive region for different queries may differ significantly according to the visualization of pretrained ViT~\cite{vit} and DETR~\cite{detr}. 
As we achieve the goal of query-adaptive sparsity in a coarse-to-fine manner, it shares some similarities with quad-tree attention~\cite{tang2022quadtree}.
Different from quad-tree attention, the goal of our bi-level routing attention is to locate a few most relevant key-value pairs, while quad-tree attention builds a token pyramid and assembles messages from all levels of different granularities.
In addition, the quad-tree requires deep recursion to cover the whole feature map, which hurts parallelism, while our bi-level routing attention can be more efficiently implemented by key/value token gathering, followed by dense matrix multiplications.
As a result, quad-tree transformer is much slower than our BiFormer.

\section{Our Approach: BiFormer}

This section elaborates the proposed approach. We start by briefly summarizing the attention mechanism in Section~\ref{sec:preliminaries}. We then introduce our novel bi-level routing attention (BRA) mechanism, which enables dynamic and query-adaptive sparsity, in Section~\ref{sec:bra}. We further show that BRA can achieve $O((HW)^\frac{4}{3})$ complexity with a proper region partition size in Section~\ref{sec:complexity}. Finally, using BRA as the core building block, we present a new hierarchical vision transformer, named BiFormer, in Section~\ref{sec:biformer}.

\subsection{Preliminaries: Attention}
\label{sec:preliminaries}

Taking queries $\mathbf{Q} \in \mathbb{R}^{N_q \times C}$, keys $\mathbf{K} \in \mathbb{R}^{N_{kv} \times C}$, and values $\mathbf{V} \in \mathbb{R}^{N_{kv} \times C}$ as input, an attention function transforms each query as a weighted sum of values, where the weights are computed as normalized dot products between the query and corresponding keys. It can be formally defined in a compact matrix form, as:
\begin{equation}
    \mathrm{Attention}(\mathbf{Q}, \mathbf{K}, \mathbf{V}) = \mathrm{softmax}\left(\frac{\mathbf{Q}\mathbf{K}^T}{\sqrt{C}}\right)\mathbf{V}.
\end{equation}
Here, the scalar factor $\sqrt{C}$ is introduced to avoid concentrated weights and gradient vanishing~\cite{vaswani2017attention}.

In transformers, the de facto building block used is multi-head self-attention (MHSA). By ``self-attention'', it means that queries $\mathbf{Q}$, keys $\mathbf{K}$ and values $\mathbf{V}$ are derived as linear projections of the same input $\mathbf{X} \in \mathbb{R}^{N \times C}$. (For vision transformers, $\mathbf{X}$ is a spatially flattened feature map, \ie, $N=H \times W$, where $H$ and $W$ are the height and width, respectively, of the feature map.) As for ``multi-head'', it implies splitting the output into $h$ chunks (\ie, heads) along the channel dimension with each chunk using an independent group of projection weights. Formally,
\begin{equation}
\begin{aligned}
    &\mathrm{MHSA}(\mathbf{X}) = \mathrm{Concat}(\mathbf{head}_0, \mathbf{head}_1, ..., \mathbf{head}_{h})\mathbf{W^o}, \\
    &\mathbf{head}_i = \mathrm{Attention}(\mathbf{X} \mathbf{W}^q_{i}, \mathbf{X} \mathbf{W}^k_i, \mathbf{X} \mathbf{W}^v_i), \\
\end{aligned}
\end{equation}
where $\mathbf{head}_i \in \mathbb{R}^{N \times \frac{C}{h}}$ is the output of the $i^{th}$ attention head. $\mathbf{W}^q_{i}, \mathbf{W}^k_i, \mathbf{W}^v_i \in \mathbb{R}^{C \times \frac{C}{h}}$ are corresponding input projection weights. An extra linear transformation with weight matrix $\mathbf{W}^o \in \mathbb{R}^{C \times C}$ is used to compose all heads.

MHSA has a complexity of $O(N^2)$, as there are $N$ queries and each query will attend to $N$ key-value pairs. Such a high complexity causes severe scalability issues \wrt the spatial resolution of the inputs.

\subsection{Bi-Level Routing Attention (BRA)}
\label{sec:bra}

\begin{algorithm}[t]
\caption{Pseudocode of BRA in a PyTorch-like style.}
\label{alg:code}
\algcomment{\fontsize{7.2pt}{0em}\selectfont \texttt{bmm}: batch matrix multiplication; \texttt{mm}: matrix multiplication. \texttt{dwconv}: depthwise convolution.
}
\definecolor{codeblue}{rgb}{0.25,0.5,0.5}
\lstset{
  backgroundcolor=\color{white},
  basicstyle=\fontsize{7.2pt}{7.2pt}\ttfamily\selectfont,
  columns=fullflexible,
  breaklines=true,
  captionpos=b,
  commentstyle=\fontsize{7.2pt}{7.2pt}\color{codeblue},
  keywordstyle=\fontsize{7.2pt}{7.2pt},
}
\begin{lstlisting}[language=python]
# input: features (H, W, C). Assume H==W.
# output: features (H, W, C).
# S: square root of number of regions.
# k: number of regions to attend.

# patchify input (H, W, C) -> (S^2, HW/S^2, C)
x = patchify(input, patch_size=H//S)

# linear projection of query, key, value
query, key, value = linear_qkv(x).chunk(3, dim=-1) 

# regional query and key (S^2, C)
query_r, key_r = query.mean(dim=1), key.mean(dim=1)

# adjacency matrix for regional graph (S^2, S^2)
A_r = mm(query_r, key_r.transpose(-1, -2))

# compute index matrix of routed regions (S^2, K)
I_r = topk(A_r, k).index

# gather key-value pairs
key_g = gather(key, I_r) # (S^2, kHW/S^2, C)
value_g = gather(value, I_r) # (S^2, kHW/S^2, C)

# token-to-token attention
A = bmm(query, key_g.transpose(-2, -1))
A = softmax(A, dim=-1)
output = bmm(A, value_g) + dwconv(value)

# recover to (H, W, C) shape
output = unpatchify(output, patch_size=H//S)

\end{lstlisting}
\end{algorithm}

To mitigate the scalability issue of MHSA, several works~\cite{swin,cswin,wang2021crossformer,tu2022maxvit,xia2022dat} propose different sparse attention mechanisms, in which each query attends to only a small number of key-value pairs instead of all. However, these existing works either use handcrafted static patterns or share the sampled subset of key-value pairs among all queries, as shown in Figure~\ref{fig:attn_comparison}. 
In this work, we explore a dynamic, query-aware sparse attention mechanism. Our key idea is to filter out most irrelevant key-value pairs in a coarse region level so that only a small portion of routed regions remain. We then apply fine-grained token-to-token attention in the union of these routed regions.
To simplify the notations, we discuss the case of single-head self-attention with a single input, although we use multi-head self-attention~\cite{vaswani2017attention} with batched input in practice. The whole algorithm is summarized with Pytorch-like~\cite{paszke2019pytorch} pseudo code in Algorithm~\ref{alg:code}. We give a detailed explanation as follows.

\bigbreak
\noindent
\textbf{Region partition and input projection}. Given a 2D input feature map $\mathbf{X} \in \mathbb{R}^{H \times W \times C}$, we start by dividing it into $S \times S$ non-overlapped regions such that each region contains $\frac{HW}{S^2}$ feature vectors. This step is done by reshaping $\mathbf{X}$ as $\mathbf{X}^r \in \mathbb{R}^{S^2 \times \frac{HW}{S^2} \times C}$. We then derive the query, key, value tensor, $\mathbf{Q}, \mathbf{K}, \mathbf{V} \in \mathbb{R}^{S^2 \times \frac{HW}{S^2} \times C}$, with linear projections:
\begin{equation}
    \mathbf{Q} = \mathbf{X}^r \mathbf{W}^q, \; \;
    \mathbf{K} = \mathbf{X}^r \mathbf{W}^k, \; \;
    \mathbf{V} = \mathbf{X}^r \mathbf{W}^v,
\end{equation}
where $\mathbf{W}^q, \mathbf{W}^k, \mathbf{W}^v \in \mathbb{R}^{C \times C}$ are projection weights for the query, key, value, respectively.

\begin{figure}[!t]
    \centering
    \includegraphics[width=.98\linewidth]{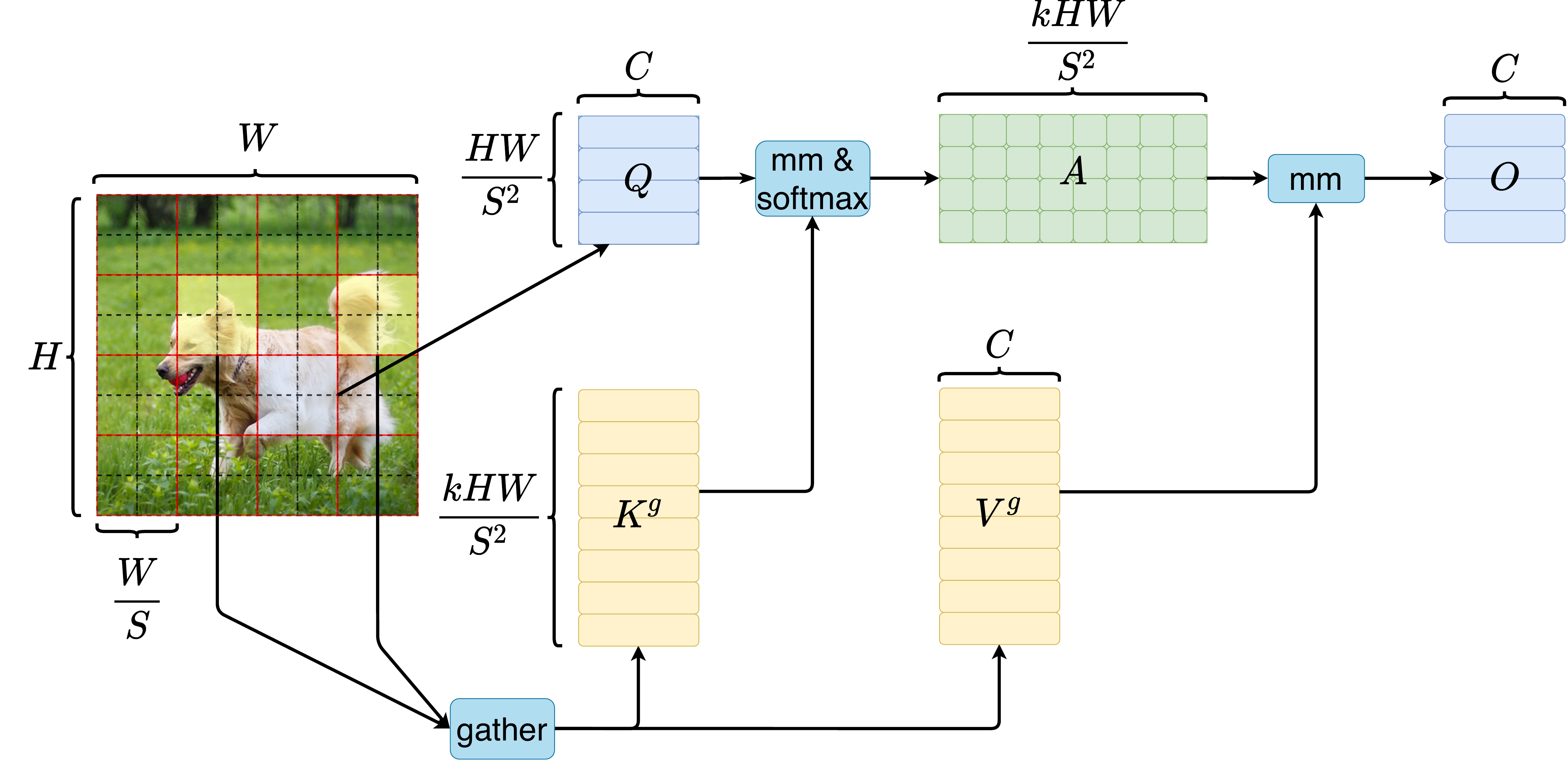}    \caption{
    By gathering key-value pairs in top $k$ related windows, we utilize the sparsity to skip computations in the most irrelevant regions, while only GPU-friendly dense matrix multiplications are involved.}
    \label{fig:fqa}
\end{figure}

\bigbreak
\noindent
\textbf{Region-to-region routing with directed graph.}
We then find the attending relationship (\ie, the regions that should be attended for each given region) by constructing a directed graph.
Specifically, we first derive region-level queries and keys, $\mathbf{Q}^r, \mathbf{K}^r \in \mathbb{R}^{S^2 \times C}$, via applying per-region average on $\mathbf{Q}$ and $\mathbf{K}$, respectively.
We then derive the adjacency matrix, $\mathbf{A}^r \in \mathbb{R}^{S^2 \times S^2}$, of region-to-region affinity graph via matrix multiplication between $\mathbf{Q}^r$ and transposed $\mathbf{K}^r$:
\begin{equation}
    \mathbf{A}^r = \mathbf{Q}^r (\mathbf{K}^r)^T. 
\end{equation}
Entries in the adjacency matrix, $\mathbf{A}_r$, measure how much two regions are semantically related. The core step that we perform next is to prune the affinity graph by keeping only top-$k$ connections for each region. Specifically, we derive a routing index matrix, $\mathbf{I}_r \in \mathbb{N}^{S^2 \times k}$, with the row-wise topk operator: 
\begin{equation}
    \mathbf{I}^r = \mathrm{topkIndex}(\mathbf{A}^r).
\end{equation}
Hence, the $i^{th}$ row of $\mathbf{I}^r$ contains $k$ indices of most relevant regions for the $i^{th}$ region. 

\bigbreak
\noindent
\textbf{Token-to-token attention.} With the region-to-region routing index matrix $\mathbf{I}^r$, we can then apply fine-grained token-to-token attention. For each query token in region $i$, it will attend to all key-value pairs residing in the union of $k$ routed regions indexed with $\mathbf{I}^r_{(i, 1)}, \mathbf{I}^r_{(i, 2)}, ..., \mathbf{I}^r_{(i, k)}$. However, it is non-trivial to implement this step efficiently, as these routed regions are expected to be scattered over the whole feature map, while modern GPUs rely on coalesced memory operations that load blocks of dozens of contiguous bytes at once. We thus gather key and value tensor first, \ie,
\begin{equation}
    \mathbf{K}^{g} = \mathrm{gather}(\mathbf{K}, \mathbf{I}^r), \; \; \mathbf{V}^g = \mathrm{gather}(\mathbf{V}, \mathbf{I}^r),
\end{equation}
where $\mathbf{K}^g, \mathbf{V}^g \in \mathbb{R}^{S^2 \times \frac{kHW}{S^2} \times C}$ are gathered key and value tensor. We can then apply attention on the gathered key-value pairs as: 
\begin{equation}
   \mathbf{O} = \mathrm{Attention}(\mathbf{Q}, \mathbf{K}^g, \mathbf{V}^g) + \mathrm{LCE}(\mathbf{V}).
\end{equation}
Here, we introduce a local context enhancement term $\mathrm{LCE}((\mathbf{V})$ as in~\cite{ren2022shunted}. Function $\mathrm{LCE}(\cdot)$ is parametrized with a depth-wise convolution, and we set the kernel size to 5.

\begin{figure*}[!t]
    \centering
    \includegraphics[width=.95\linewidth]{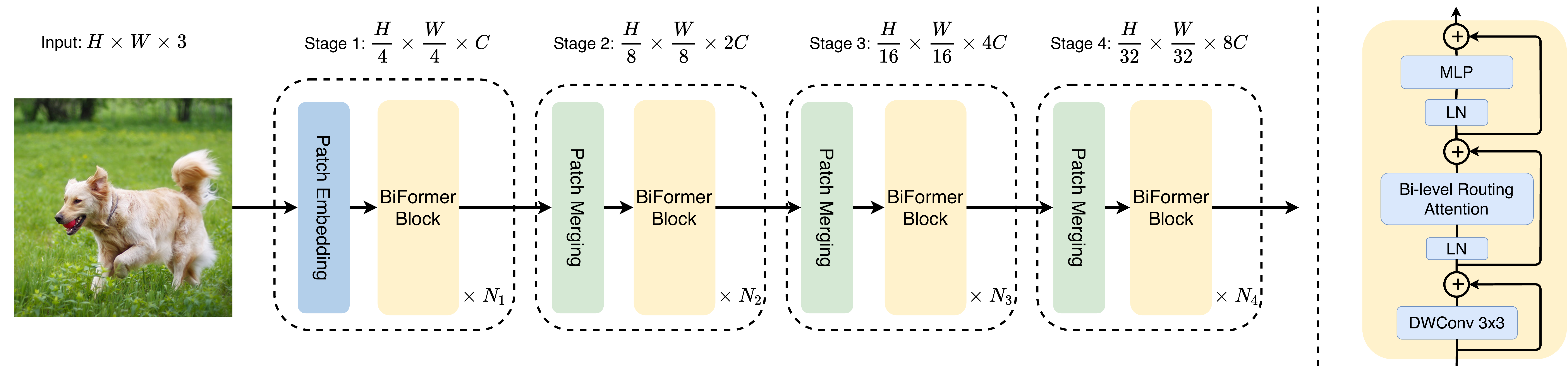}
    \vspace{-3mm}
    \caption{\textbf{Left}: The overall architecture of our BiFormer. Refer to Table~\ref{table:config} for configurations. \textbf{Right}: Details of a BiFormer Block.}
    \label{fig:biformer}
    \vspace{-1em}
\end{figure*}

\subsection{Complexity Analysis of BRA}
\label{sec:complexity}

The proposed bi-level routing attention enables direct long-range dependency modeling similar to vanilla attention. However, we show here that BRA has a much lower complexity of $O((HW)^\frac{4}{3})$ with a proper region partition factor $S$ compared to vanilla attention, which has a complexity of $O((HW)^2)$, and to quasi-global axial attention~\cite{cswin,huang2019ccnet}, which has a complexity of $O((HW)^\frac{3}{2})$.

The computation of BRA consists of three parts: linear projection, region-to-region routing, and token-to-token attention. The total amount of computations is therefore:
\begin{equation}\label{eq:complexity}
\begin{aligned}
    \mathrm{FLOPs} &= \mathrm{FLOPs}_{proj} + \mathrm{FLOPs}_{routing} + \mathrm{FLOPs}_{attn} \\
          &= 3HWC^2 + 2 (S^2)^2 C + 2HW k \frac{HW}{S^2} C \\
          &= 3HWC^2 + C (2S^4 + \frac{k(HW)^2}{S^2} + \frac{k(HW)^2}{S^2}) \\
          &\geq 3HWC^2 + 3C(2S^4 \cdot \frac{k(HW)^2}{S^2} \cdot \frac{k(HW)^2}{S^2})^\frac{1}{3} \\
          &= 3HWC^2 + 3Ck^\frac{2}{3}(2HW)^\frac{4}{3},
\end{aligned}
\end{equation}
where $C$ is the token embedding dimension (\ie, number of channels of the feature map), and $k$ is the number of regions to attend (``$k$'' in ``top-$k$'').
Here, the inequality of arithmetic and geometric means has been applied.
The equality in Eq.~\ref{eq:complexity} holds if and only if $2S^4 = \frac{k(HW)^2}{S^2}$. Therefore:
\begin{equation}\label{eq:condition}
    S = (\frac{k}{2}(HW)^2)^\frac{1}{6}.
\end{equation}
In other words, BRA achieves $O((HW)^\frac{4}{3})$ complexity if we scale the region partition factor $S$ \wrt the input resolution according to Eq.~\ref{eq:condition}.

\subsection{Architecture Design of BiFormer}
\label{sec:biformer}

Using BRA as a basic building block, we propose a new general vision transformer, BiFormer. As shown in Figure~\ref{fig:biformer}, we follow the recent state-of-the-art vision transformers~\cite{swin,cswin,tu2022maxvit} to use a four-stage pyramid structure.
Specifically, in stage $i$, we use an overlapped patch embedding in the first stage and a patch merging module~\cite{li2022uniformer,ren2022shunted} in the second to fourth stages to reduce the input spatial resolution while increasing the number of channels, followed by $N_i$ consecutive BiFormer blocks to transform the features. In each BiFormer block, we follow recent works~\cite{li2022uniformer,tu2022maxvit,chu2021twins} to use a $3 \times 3$ depthwise convolution at the beginning to encode relative position information implicitly. We then apply a BRA module and 2-layer MLP module with expansion ratio $e$ sequentially for cross-location relation modeling and per-location embedding, respectively.

We instantiate BiFormer with 3 different model sizes by scaling the network width (\ie, the number of base channels $C$) and depth (\ie, the number of BiFormer blocks used in each stage, $N_i, i=1, 2, 3, 4$), as listed in Table~\ref{table:config}. They share other configurations. We set each attention head to 32 channels, and MLP expansion ratio $e$=3. For BRA, we use $topk=1, 4, 16, S^2$\footnote{In the final stage, $topk=S^2$ means that we use full self-attention.} for the 4 stages, and region partition factor $S=7/8/16$ for classification/semantic segmentation/object detection task, due to different input resolutions. 

\begin{table}[htpb]
\begin{center}
  \resizebox{0.95\linewidth}{!}{
  \begin{tabular}{l|c|c|c|c}
    \toprule
     Models & \#Channels. & \#Blocks & Params & FLOPs\\
    \midrule
    BiFormer-T & 64 & [2, 2, 8, 2]  & 13M & 2.2G  \\
    BiFormer-S & 64 & [4, 4, 18, 4] & 26M & 4.5G \\
    BiFomrer-B & 96 & [4, 4, 18, 4] & 57M & 9.8G \\
    \bottomrule
    \end{tabular}
    }
\end{center}
\vspace{-1em}
\caption{Network width and depth of different model variants. 
The FLOPs are calculated with 224 $\times$ 224 input.}
\label{table:config}
\end{table}

\begin{table}[htpb]
\begin{center}
  \resizebox{0.95\linewidth}{!}{
  \begin{tabular}{l|c|c|c}
    \toprule
    \multirow{2}{*}{Model} & FLOPs & Params & Top-1 Acc.  \\
        & (G) & (M) & (\%) \\
    \midrule
    ResNet-18~\cite{he2016resnet} & 1.8 & 11.7 & 69.8 \\
    RegNetY-1.6G~\cite{radosavovic2020regnet} & 1.6 & 11.2 & 78.0 \\
    PVTv2-b1~\cite{wang2022pvtv2} & 2.1 & 13.1 & 78.7 \\
    Shunted-T~\cite{ren2022shunted} & 2.1 & 11.5 & 79.8 \\
    QuadTree-B-b1~\cite{tang2022quadtree} & 2.3 & 13.6 & 80.0 \\
    \rowcolor{LightCyan}
    BiFormer-T & 2.2 & 13.1 & \textbf{81.4} \\
    \midrule
    Swin-T~\cite{swin} & 4.5  & 29 & 81.3 \\
    CSWin-T~\cite{cswin} & 4.5 & 23 & 82.7 \\
    DAT-T~\cite{xia2022dat} & 4.6 & 29 & 82.0 \\
    CrossFormer-S~\cite{wang2021crossformer} & 5.3 & 31 & 82.5  \\
    RegionViT-S~\cite{chen2021regionvit} & 5.3 & 31 & 82.6 \\
    QuadTree-B-b2~\cite{tang2022quadtree} & 4.5 & 24 & 82.7 \\
    MaxViT-T~\cite{tu2022maxvit} & 5.6 & 31 & 83.6 \\
    ScalableViT-S~\cite{ScalableViT} & 4.2 & 32 & 83.1 \\
    Uniformer-S* & 4.2 & 24 & 83.4 \\
    Wave-ViT-S*~\cite{yao2022wavevit} & 4.7 & 23 & 83.9 \\
    \rowcolor{LightCyan}
    BiFormer-S & 4.5 & 26 & \textbf{83.8}\\
    \rowcolor{LightCyan}
    BiFormer-S* & 4.5 & 26 & \textbf{84.3} \\
    \midrule
    Swin-B~\cite{swin} & 15.4 & 88 & 83.5 \\
    CSWin-B~\cite{cswin} & 15.0 & 78 & 84.2 \\
    CrossFormer-L~\cite{wang2021crossformer} & 16.1 & 92 & 84.0 \\
    ScalableViT-B~\cite{ScalableViT} & 8.6 & 81 & 84.1 \\
    Uniformer-B*~\cite{li2022uniformer} & 8.3 & 50 & 85.1 \\
    Wave-ViT-B*~\cite{yao2022wavevit} & 7.2 & 34 & 84.8 \\
    \rowcolor{LightCyan}
    BiFormer-B & 9.8 & 57 & \textbf{84.3} \\
    \rowcolor{LightCyan}
    BiFormer-B* & 9.8 & 58 & \textbf{85.4} \\
    \bottomrule
    \end{tabular}
    }
\end{center}
\vspace{-2mm}
\caption{Comparison of different backbones on ImageNet-1K. All models are trained and evaluated on images of resolution $224 \times 224$. ``*'' indicates that the model is trained with token labeling~\cite{jiang2021token_labeling}. Methods are grouped by the amount of computations.
}
\label{table:quantitative}
\vspace{-4mm}
\end{table}

\section{Experiments}

We evaluate the effectiveness of our proposed BiFormer experimentally on a series of mainstream computer vision tasks including image classification (Sec.~\ref{sec:classification}), object detection and instance segmentation (Sec.~\ref{sec:detection}), and semantic segmentation (Sec.~\ref{sec:segmentation}). Specifically, we train from scratch on ImageNet1K~\cite{deng2009imagenet} for image classification. We then fine-tune the pretrained backbones on COCO~\cite{lin2014coco} for object detection and instance segmentation, and on ADE20K~\cite{zhou2019ade20k} for semantic segmentation. Additionally, we conduct ablation study to verify the effectiveness of the proposed bi-level routing attention and other architecture design choices of BiFormer in Sec.~\ref{sec:ablation}. Finally, to verify query-adaptive, sparse patterns are achieved by bi-level routing attention, we visualize the attention map in Sec.~\ref{sec:visualization}.

\subsection{Image Classification on ImageNet-1K}
\label{sec:classification}

\begin{table*}[htpb]
\begin{center}
  \resizebox{0.95\linewidth}{!}{
  \begin{tabular}{l|cccccc|cccccc}
    \toprule
    \multirow{2}{*}{Backbone} & \multicolumn{6}{c|}{RetinaNet 1× schedule} & \multicolumn{6}{c}{Mask R-CNN 1× schedule}  \\
     &  $mAP$ & $AP_{50}$  & $AP_{75}$ & $AP_S$ &  $AP_M$ & $AP_L$ & $mAP^b$ & $AP^b_{50}$  & $AP^b_{75}$ & $mAP^m$ &  $AP^m_{50}$ & $AP^m_{75}$ \\
    \midrule
    Swin-T~\cite{swin} & 41.5 & 62.1 & 44.2 & 25.1 & 44.9 & 55.5 & 42.2 & 64.6 & 46.2 & 39.1 & 61.6 & 42.0 \\
    DAT-T~\cite{xia2022dat} & 42.8 & 64.4 & 45.2 & 28.0 & 45.8 & 57.8 & 44.4 & 67.6 & 48.5 & 40.4 & 64.2 & 43.1 \\
    CSWin-T~\cite{cswin} & - & - & - & - & - & - & 46.7 & 68.6 & 51.3 & 42.2 & 65.6 & 45.4 \\
    CrossFormer-S~\cite{wang2021crossformer} & 44.4 & 55.3 & 38.6 & 19.3 & 40.0 & 48.8 & 45.4 & 68.0 & 49.7 & 41.4 & 64.8 & 44.6 \\
    QuadTree-B2~\cite{tang2022quadtree} & 46.2 & 67.2 & 49.5 & 29.0 & 50.1 & 61.8 & - & - & - & - & - & - \\
    WaveViT-S*~\cite{yao2022wavevit} & 45.8 & 67.0 & 49.4 & 29.2 & 50.0 & 60.8 & 46.6 & 68.7 & 51.2 & 42.4 & 65.5 & 45.8 \\
    \rowcolor{LightCyan}
    BiFormer-S & 45.9 & 66.9 & 49.4 & \textbf{30.2} & 49.6 & 61.7 & \textbf{47.8} & \textbf{69.8} & \textbf{52.3} & \textbf{43.2} & \textbf{66.8} & \textbf{46.5} \\
    \midrule
    \midrule
    Swin-S~\cite{swin} & 44.5 & 65.7 & 47.5 & 27.4 & 48.0 & 59.9 & 44.8 & 66.6 & 48.9 & 40.9 & 63.4 & 44.2 \\
    DAT-S~\cite{xia2022dat} & 45.7 & 67.7 & 48.5 & 30.5 & 49.3 & 61.3 & 47.1 & 69.9 & 51.5 & 42.5 & 66.7 & 45.4 \\
    CSWin-S~\cite{cswin} & - & - & - & - & - & - & 47.9 & 70.1 & 52.6 & 43.2 & 67.1 & 46.2 \\
    CrossFormer-B~\cite{wang2021crossformer} & 46.2 & 67.8 & 49.5 & 30.1 & 49.9 & 61.8 & 47.2 & 69.9 & 51.8 & 42.7 & 66.6 & 46.2 \\
    QuadTree-B3~\cite{tang2022quadtree} & 47.3 & 68.2 & 50.6 & 30.4 & 51.3 & 62.9 & - & - & - & - & - & - \\
    Wave-ViT-B*~\cite{yao2022wavevit} & 47.2 & 68.2 & 50.9 & 29.7 & 51.4 & 62.3 & 47.6 & 69.1 & 52.4 & 43.0 & 66.4 & 46.0 \\
    \rowcolor{LightCyan}
    BiFormer-B & 47.1 & \textbf{68.5} & 50.4 & \textbf{31.3} & 50.8 & 62.6 & \textbf{48.6} & \textbf{70.5} & \textbf{53.8} & \textbf{43.7} & \textbf{67.6} & \textbf{47.1} \\    
    \bottomrule
    \end{tabular}
    }
\end{center}
\vspace{-1em}
\caption{Comparison based on the object detection (left group) and instance segmentation (right group) tasks, on the COCO 2017 dataset.
%
}
\label{table:quantitative_det}
\vspace{-1em}
\end{table*}

\noindent
\textbf{Settings}. We conduct image classification experiments on the ImageNet-1K~\cite{deng2009imagenet} dataset, following the experimental settings of DeiT~\cite{touvron2021deit} for fair comparison. Specifically, each model is trained 300 epochs with input size of $224 \times 224$.  We take AdamW as the optimizer with weight decay of 0.05, and apply cosine decay learning rate schedule with an initial learning rate of 0.001, while the first 5 epochs are utilized for linear warm-up~\cite{goyal2017accurate_warmup}. The batch size is set to 1024. To avoid overfitting, we apply regularization techniques including RandAugment~\cite{cubuk2020randaugment} (rand-m9-mstd0.5-inc1), MixUp~\cite{zhang2017mixup} ($prob=0.8$), CutMix~\cite{yun2019cutmix} ($prob=1.0$), Random Erasing ($prob=0.25$), and increasing stochastic depth~\cite{huang2016stochastic_depth} ($prob=0.1/0.15/0.4$ for BiFormer-T/S/B, respectively). To fairly compare the models trained with token labeling~\cite{jiang2021token_labeling}, including Uniformer~\cite{li2022uniformer} and WaveViT~\cite{yao2022wavevit}, we also provide a version trained with the same recipe provided by WaveViT.

\noindent
\textbf{Results.} We compare our method with several closely related methods and/or recent state-of-the-arts. Quantitative results are listed in Table~\ref{table:quantitative}, where models are grouped by the amount of computations (FLOPs).
In all 3 groups, our model consistently outperforms other compared ones.
For example, for models in the smallest group ($\sim$2G FLOPs), our BiFormer-T achieves 81.4\% top-1 accuracy, 1.4\% better than the most competitive QuadTree-b1~\cite{tang2022quadtree}.
For models in the second group ($\sim$4G FLOPs), BiFormer-S achieves 83.8\% top-1 accuracy. To the best of our knowledge, this is the best result without extra training data or training tricks. In addition, using the distillation technique named token labeling~\cite{jiang2021token_labeling}, the accuracy of BiFormer-S can be further boosted to 84.3\%, which implies that there is a huge potential for the proposed architecture.
For models in the largest group ($\sim$10G FLOPs), BiFormer-B achieves an even better performance than those of existing models with the amount of computations reaching up to $\sim$16G FLOPs, such as Swin-B~\cite{swin}, CSWin-B~\cite{cswin} and CrossFormer-L~\cite{wang2021crossformer}.

\begin{table}[htpb]
\begin{center}
  \resizebox{0.95\linewidth}{!}{
  \begin{tabular}{l|c|cc}
    \toprule
    \multirow{2}{*}{Backbone} & S-FPN & \multicolumn{2}{c}{Upernet}  \\
            & mIoU(\%) & mIoU(\%) & MS mIOU(\%) \\
    \midrule
    Swin-T~\cite{swin} & 41.5 & 44.5 & 45.8 \\
    DAT-T~\cite{xia2022dat} & 42.6 & 45.5 & 46.4 \\
    CSWin-T~\cite{cswin} & 48.2 & 49.3 & 50.7\\
    CrossFormer-S~\cite{wang2021crossformer} & 46.0 & 47.6 & 48.4 \\
    Shunted-S~\cite{ren2022shunted} & 48.2 & 48.9 & 49.9 \\
    WaveViT-S*~\cite{yao2022wavevit} & - & - & 49.6 \\
    \rowcolor{LightCyan}
    BiFormer-S & \textbf{48.9} & \textbf{49.8} &  \textbf{50.8} \\
    \midrule
    \midrule
    Swin-S~\cite{swin} & - & 47.6 & 49.5 \\
    DAT-S~\cite{xia2022dat} & 46.1 & 48.3 & 49.8 \\
    CSWin-S~\cite{cswin} & 49.2 & 50.4 & 51.5 \\
    CrossFormer-B~\cite{wang2021crossformer} & 47.7 & 49.7 & 50.6 \\
    Uniformer-B~\cite{li2022uniformer} & 48.0 & 50.0 & 50.8 \\
    WaveViT-B*~\cite{yao2022wavevit} & - & - & 51.5 \\ 
    \rowcolor{LightCyan}
    BiFormer-B & \textbf{49.9} & \textbf{51.0} &  \textbf{51.7} \\
    \bottomrule
    \end{tabular}
    }
\end{center}
\vspace{-4mm}
\caption{Comparison based on semantic segmentation with two segmentation heads (Semantic FPN and UpperNet), on ADE20K.
}
\vspace{-2mm}
\label{table:quantitative_seg}
\end{table}

\subsection{Object Detection and Instance Segmentation}
\label{sec:detection}

\noindent
\textbf{Settings}. We evaluate the models for object detection and instance segmentation on COCO 2017~\cite{lin2014coco}.
For a fair comparison, all experiments are conducted with the MMDetection~\cite{chen2019mmdetection} toolbox. RetinaNet~\cite{lin2017retinanet} and Mask R-CNN~\cite{he2017maskrcnn} frameworks are used for object detection and instance segmentation, respectively.
Before training on COCO, we initialize the backbone with weights pretrained on ImageNet-1K, while leaving all other layers randomly initialized.
The models are trained with the standard $1\times$ schedule (12 epochs) provided by MMDetection, except that we use the AdamW optimizer~\cite{loshchilov2017adamw}, instead of SGD.
We use an initial learning rate of $1e-4$, and a batch size of 16,  while the weight decay is set as $1e-4$ and $5e-2$ for RetinaNet and Mask R-CNN, respectively.
During training, we resize the input images by fixing the shorter side to 800 pixels while keeping the longer side not exceeding 1,333 pixels.

\noindent
\textbf{Results}. We list results in Table~\ref{table:quantitative_det}. For object detection with RetinaNet,
we report mean Average Precision ($mAP$), Average Precision ($AP$) at different IoU thresholds (50\%, 75\%) and for three object sizes (\ie small, medium, and large (S/M/L)).
From the results, we can see that while the overall performance of BiFormer is only comparable to some most competitive existing methods, such as WaveViT and QuadTree-B, the performance on small objects ($AP_M$) outperforms these methods significantly. 
This may be because the BRA saves computations via sparse sampling instead of downsampling. Hence, it preserves fine-grained details, which are crucial for small objects.
For instance segmentation with Mask R-CNN, we report bounding box and mask Average Precision ($AP^b$ and $AP^m$) at different IoU thresholds (50\%, 75\%).
As shown in Table~\ref{table:quantitative_det}, our method shows a clear advantage in this task on all metrics.

\subsection{Semantic Segmentation on ADE20K}
\label{sec:segmentation}

\begin{table}[htpb]
\begin{center}
  \resizebox{0.9\linewidth}{!}{
  \begin{tabular}{l|c|c}
    \toprule
     Sparse Attention  & IN1K  & ADE20K \\
                     & Top1(\%) & mIoU(\%) \\
    \midrule
    Sliding window~\cite{ramachandran2019stand_slidingattn} & 81.4 & - \\
    Shifted window~\cite{swin} & 81.3 & 41.5 \\
    Spatially Sep~\cite{chu2021twins} & 81.5 & 42.9 \\
    Sequential Axial~\cite{ho2019seqaxial} & 81.5 & 39.8 \\
    Criss-Cross~\cite{huang2019ccnet} & 81.7 & 43.0 \\
    Cross-shaped window~\cite{cswin} & 82.2 & 43.4 \\
    Deformable~\cite{xia2022dat} & 82.0 & 42.6 \\
    Block-Grid~\cite{tu2022maxvit} & 81.8 & 42.8 \\
    \rowcolor{LightCyan}
    Bi-level Routing & \textbf{82.7} &\textbf{44.8}\\
    \bottomrule
    \end{tabular}
    }
\end{center}
\vspace{-1.5em}
\caption{Ablation study on different attention mechanisms. All models follow the architecture design of the Swin-T model.}
\label{table:ablation_attention}
\vspace{-1em}
\end{table}

\noindent
\textbf{Settings}. Following existing works,  we conduct our semantic segmentation experiments on the ADE20K~\cite{zhou2019ade20k} dataset based on MMSegmentation~\cite{contributors2020mmseg}.
We do comparisons under both Semantic FPN~\cite{kirillov2019semanticfpn} and UperNet~\cite{xiao2018upernet} frameworks.
In both cases, the backbone is initialized with ImageNet-1K pretrained weights, and other layers use random initialization. Models are optimized with the AdamW optimizer and the batch size is set as 32.
For a fair comparison, our Semantic FPN experiments use the same setting as PVT~\cite{wang2021pvt} to train the model 80k steps.
Our Upernet experiments use the same setting as Swin Transformer~\cite{swin} to train the model 160k iterations.
%

\noindent
\textbf{Results}. 
Table~\ref{table:quantitative_seg} shows the results of the two different frameworks.
It shows that with the Semantic FPN framework, our BiFormer-S/B achieves 48.9/49.9 mIoU, respectively, improving CSWin-T/S by 0.7 mIoU. A similar performance gain for the UperNet framework is also observed.

\subsection{Ablation Study}
\label{sec:ablation}

\begin{figure*}[!t]
    \centering
    \includegraphics[width=.95\linewidth]{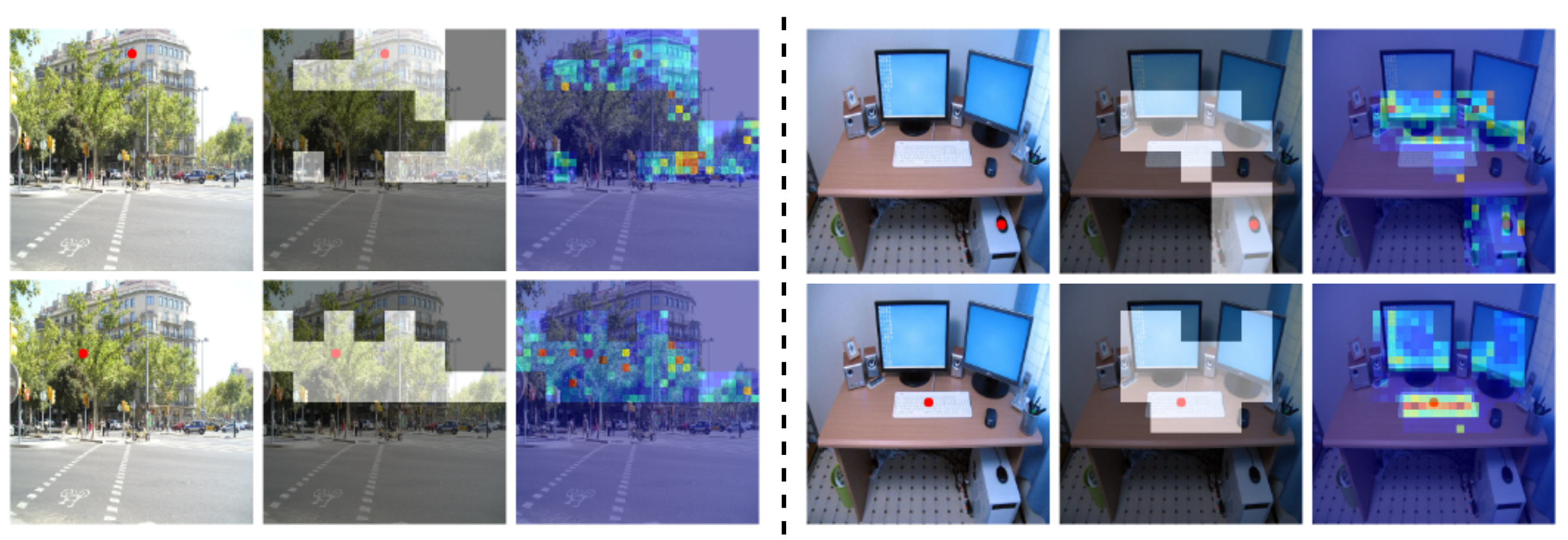}
    \vspace{-4mm}
    \caption{Visualization of the attention maps for two scenes. For each scene, we visualize two query positions on the input image (left), corresponding routed regions (middle), and a final attention heatmap (right).}
    \label{fig:attn_vis}
    \vspace{-4mm}
\end{figure*}

\noindent
\textbf{The effectiveness of BRA}.
We compare BRA with several existing sparse attention mechanisms.
Following~\cite{cswin}, we align macro architecture designs with Swin-T~\cite{swin} for a fair comparison.
Specifically, we use 2, 2, 6, 2 blocks for the four stages,  non-overlapped patch embedding, set the initial patch embedding dimension $C=96$ and  MLP expansion ratio $e=4$.
The results are reported in Table~\ref{table:ablation_attention}.
Our bi-level routing attention has significantly better performance than existing sparse attention mechanisms, in terms of both image classification and semantic segmentation.

\noindent
\textbf{Other architecture design choices.} Using the Swin-T layout as the baseline, we present a summary of other modifications that we have applied, which further boost our BiFormer-S model to state-of-the-art performances on the ImageNet-1K dataset.
These modifications include: (1) replacing non-overlapped patch embedding~\cite{swin} with overlapped one~\cite{wang2022pvtv2,cswin,ren2022shunted}, (2) using deeper layout (\ie stacking more blocks in each stage, while reducing the base channels from 96 to 64 and MLP expansion ratio from 4 to 3 to keep similar FLOPs.), (3) adding convolution position encoding~\cite{chu2021twins,li2022uniformer} at the beginning of the BiFormer blocks, and (4) applying token labeling~\cite{jiang2021token_labeling,yao2022wavevit,li2022uniformer} training technique.
As shown in Table~\ref{table:ablation_arch}, simply using a deeper layout can improve the performance significantly. However, this factor is usually not discussed in existing works.

\begin{table}[htpb]
\begin{center}
  \resizebox{0.95\linewidth}{!}{
  \begin{tabular}{l|c|c|l}
    \toprule
    Architecture design  & Params & FLOPs  & IN1K Top1 \\
                        & (M)    &  (G)   & (\%) \\
    \midrule
    Baseline (Swin-T layout) & 29 & 4.6 & 82.7 \\
    +Overlapped patch emb. & 31 & 4.9 & 82.8 \textcolor{blue}{(+0.1)}  \\
    +Deeper layout & 25 & 4.5 & 83.5 \textcolor{blue}{(+0.7)} \\
    +Convolution pos. enc. & 26 & 4.5 & 83.8 \textcolor{blue}{(+0.3)} \\
    +Token Labling & 29 & 4.9 & 84.3 \textcolor{blue}{(+0.5)} \\
    \bottomrule
    \end{tabular}
    }
\end{center}
\vspace{-4mm}
\caption{Ablation path from Swin-T~\cite{swin} layout architecture to BiFormer-S. Note that the modifications are applied sequentially.}
\vspace{-2mm}
\label{table:ablation_arch}
\end{table}

\subsection{Visualization of Attention Map}
\label{sec:visualization}

To further understand how bi-level routing attention works, we visualize routed regions and attention response \wrt query positions.
For this visualization, we use the routing indices and attention scores extracted from the final BiFormer block of the $3^{rd}$ stage, which is the major stage consuming most computations.
We demonstrate two scenes in Figure~\ref{fig:attn_vis}.
In both cases, we can clearly observe that semantically related regions are successfully located.
For example, in the first scene, which is a street view, if the query position is on a building or a tree, the corresponding routed regions cover the same or similar entities.
In the second indoor scene, when we place the query position on the mouse, the routed regions contain part of the host, keyboard, and display, even though these regions are not adjacent to each other. This implies that our bi-level routing attention can capture long-range inter-object relationships.

\section{Limitation and Future Work}

Compared to sparse attention with simple static patterns, we introduce an extra step to locate the regions to attend, where we build and prune a region-level graph and gather key-value pairs from the routed regions.
While this step does not incur much computation as it operates at a coarse region level, it inevitably incurs extra GPU kernel launch and memory transactions. Hence, BiFormer has lower throughput than some existing models with similar FLOPs on GPU due to overheads of kernel launch and memory bottleneck.
Nonetheless, this problem can be mitigated via engineering efforts, such as GPU kernel fusion.
We will explore efficient sparse attention and vision transformer with hardware awareness in our future works.

\section{Conclusion}

We propose bi-level routing attention to enable efficient allocation of computations in a dynamic, query-aware manner.
The core idea of BRA is to filter out the most irrelevant key-value pairs at a coarse region level.
It is achieved by first building and pruning a region-level directed graph, and then applying fine-grained token-to-token attention in the union of routed regions. 
We have analyzed the computational complexity of BRA and demonstrated that it achieves $O((HW)^\frac{4}{3})$ with a proper region partition size.
Using BRA as the core building block, we propose BiFormer, a new vision transformer that has shown superior performances on four popular vision tasks, image classification, object detection, instance segmentation, and semantic segmentation.

\clearpage

\appendix
\section*{\Large{Appendix}}

\section{Discussion on Regional Representations}

In our proposed bi-level routing attention, we derive the regional representations ($\mathbf{Q}^r$ and $\mathbf{K}^r$) with average pooling for region-to-region routing. We justify the choice here.

In fact, as the goal of region-to-region routing is to find the most related tokens for token-to-token attention in the next step, it is reasonable to maximize \emph{the average token-to-token affinity scores between the two regions}.
However, this is equivalent to maximizing \emph{the affinity score between the average tokens of the two regions}, because
\begin{equation}\label{eq:region_pooling}
    \frac{1}{|\Omega| \cdot |\Omega^\prime|} \sum_{i \in \Omega} \sum_{j \in \Omega^\prime} \mathbf{Q}_i \mathbf{K}_j =  \frac{ \sum_{i \in \Omega} \mathbf{Q}_i }{|\Omega|} \cdot \frac{ \sum_{j \in \Omega^\prime} \mathbf{K}_j }{|\Omega^\prime|},
\end{equation}
where we denote the set of token indices of the two regions with $\Omega$ and $\Omega^\prime$. 

\section{Throughput Comparison}

To demonstrate the computation efficiency of the proposed bi-level routing attention, we compare the throughputs of models using different attention mechanisms.
Specifically, we replace the shift window attention modules in Swin-T~\cite{swin} with 
quad-tree attention~\cite{tang2022quadtree} modules to form  QuadTree-STL, and with our bi-level routing attention modules to form BiFormer-STL.
We then use the widely used timm~\cite{rw2019timm} script to benchmark the training and inference throughput on a 32 GB Tesla V100 GPU with a batch size of 128 and image resolution of $ 224 \times 224$.

\begin{figure}[!t]
    \centering
    \includegraphics[width=.95\linewidth]{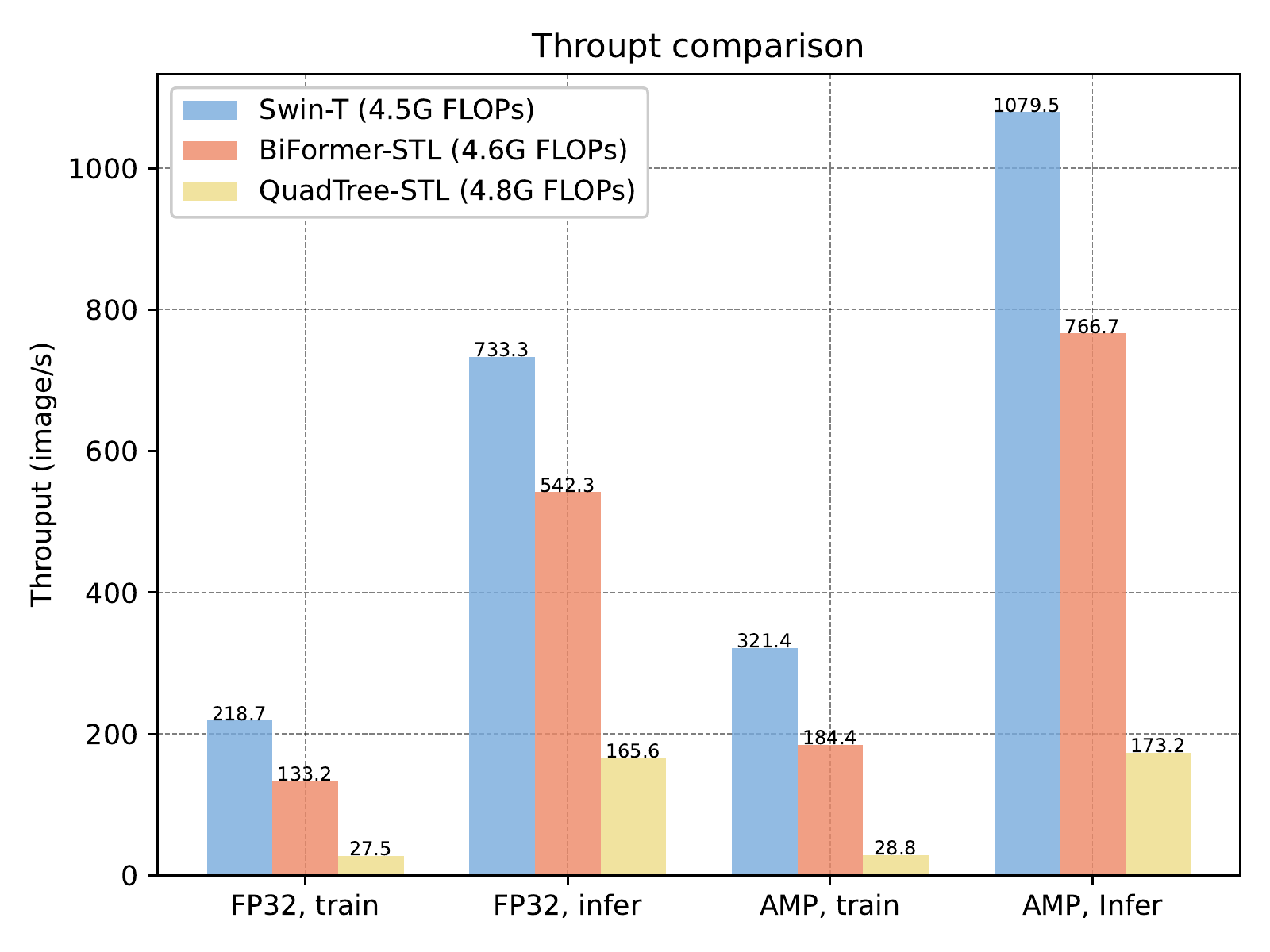}
    \caption{Throughput comparison on a 32GB Tesla V100 GPU. The suffix ``STL'' denotes \textbf{S}win-\textbf{T} \textbf{L}ayout, which means we use Swin-T~\cite{swin} backbone with only attention module being replaced.
    %
    %
    We report results under both FP32 precision and automatic mixed precision (AMP) modes.
    }
    \label{fig:throughput}
\end{figure}

As shown in Figure~\ref{fig:throughput}, Swin-T has the highest throughput due to its simplicity. 
Switching to our bi-level routing attention(BRA), the training and inference throughput of BiFormer-STL decrease by $\sim$30\% and $\sim$40\% respectively in comparison with Swin-T.
This is caused by extra GPU kernel launch and memory transactions caused by the routing process (\ie locating the regions to attend and gather key-value pairs).
Nonetheless, BiFormer-STL is still $3\times  \sim 6\times $ faster than QuadTree-STL.
This is due to that on the one hand the recursive nature of quad-tree attention hurts the parallelism, on the other hand quad-tree attention relies on sparse matrix multiplications which are inefficient on GPUs, while our BRA can be efficiently implemented with key-value token gathering followed by GPU-friendly dense matrix multiplications.

It is worth noting that, the overheads of both memory transactions and kernel launch incurred by the routing process can be reduced via engineering efforts such as GPU kernel fusion. We leave this optimization to our future work.

\begin{figure*}[htpb]
    \centering
    \includegraphics[width=.92\linewidth]{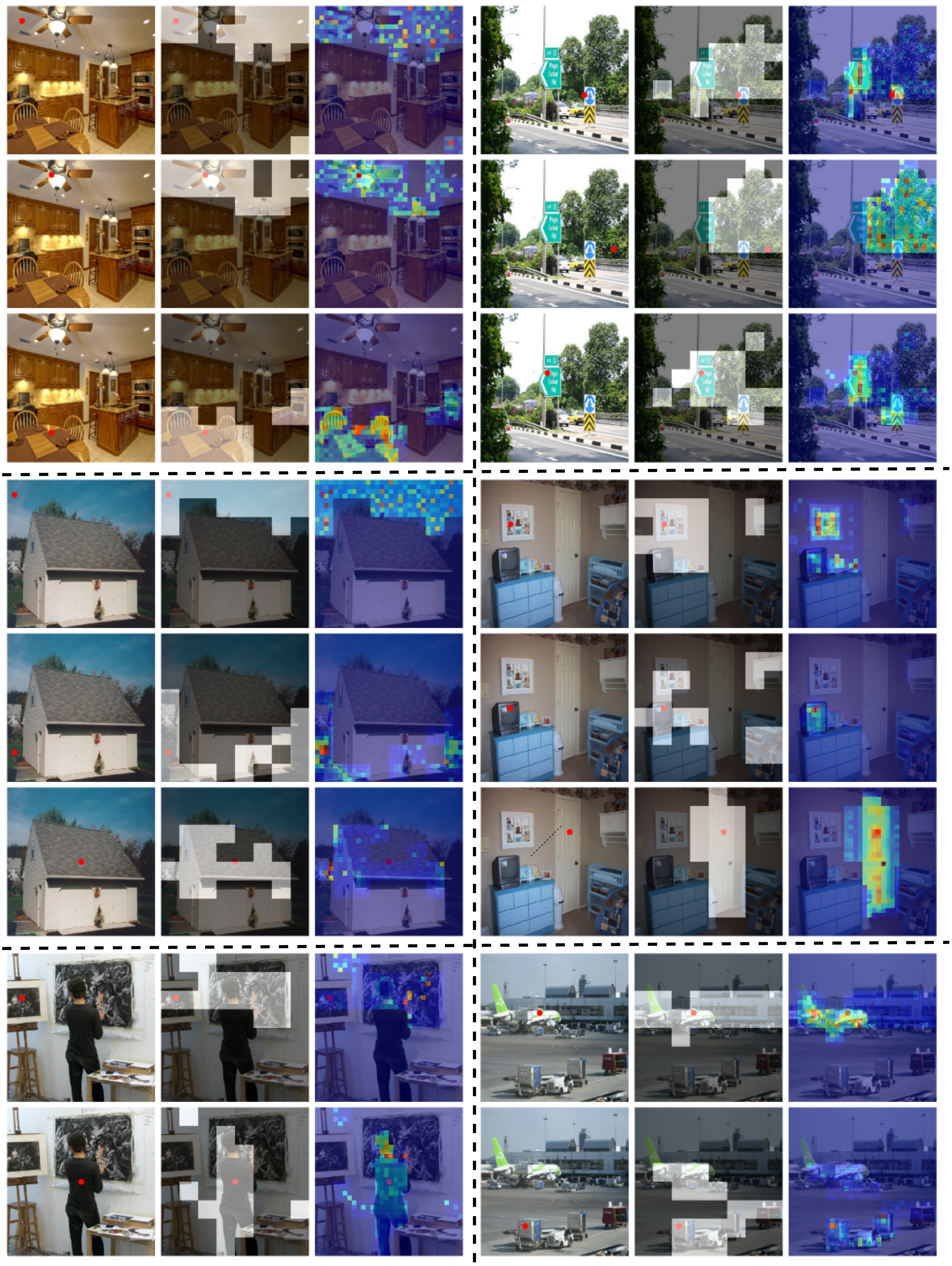}
    \caption{More attention map visualization results. For each scene, We demonstrate 2-3 query positions on the input image (left), corresponding routed regions (middle) and final attention heat map (right). }
    \label{fig:attn_vis_supp}
\end{figure*}

\section{Choices of top-$k$ and partition factor $S$}

\begin{table}[htpb]
\begin{center}
  \resizebox{0.9\linewidth}{!}{
  \begin{tabular}{c|c|c|c|c}
    \toprule
    $S$     &   $k$        & \#tokens to attend   & Acc      & im/s (FP32) \\
    \midrule
    7       &  1,4,16,49   & 64,64,64,49          & \textbf{82.7}     & 522.3        \\
    \midrule
    7       &  1,2,8,32    & 64,32,32,32          & 82.4       & 563.2      \\
    7       &  2,8,32,49   & 128,128,128,49       & 82.6     & 419.9       \\
    8,4,2,1 &  2,2,2,1     & 98,98,98,49          & 82.3     & \textbf{606.2}       \\
    \bottomrule
    \end{tabular}
    }
\end{center}
\vspace{-1em}
\caption{Ablation study on top-$k$ and partition factor $S$.}
\label{table:abl_s_k}
\end{table}

In the paper, $S$ and $k$ were chosen more with consideration of engineering issues.
\textbf{(1)} $S$ is chosen as a divisor of the training size to avoid padding, which slows down the training and may also degrade the performance.
For example, in image classification where the resolution is $224=7 \times 32$, we use $S=7$ so that it is a divisor of the size of feature maps in every stage.
This is similar to SWinTransformer~\cite{swin}, which uses a window size of 7. 
\textbf{(2)} In dense prediction tasks, we use larger $S$ to balance the complexity of region-level routing and token-level attention to achieve overall lower complexity. 
One can find hints from Eq. 9 of the paper, though we do not strictly follow the scaling rule due to the size divisor constraint.
\textbf{(3)} We gradually increase $k$ to keep a reasonable number of tokens to attend as the region size becomes smaller in later stages.

It is possible to try different combinations of $S$ and $k$. We show ablation results on IN-1K in Table~\ref{table:abl_s_k}, based on BiFormer-STL (as in the paper). A key observation from these experiments is that increasing the number of tokens to attend may even hurt the accuracy. This implies the explicit sparsity constraint may serve as a regularization to avoid distractions from the background. 

\section{Adapting Pretrained Plain ViT with BRA}

Recently, to take advantage of large-scale pretraining with masked image modeling, a new research direction emerges to adapt plain ViT~\cite{vit} for dense prediction tasks~\cite{li2022vitdet,chen2022vitadapter}.
%
%
Here we explore adapting pre-trained plain ViT~\cite{vit} for semantic segmentation with our proposed BRA.
 
Specifically, we replace all or part of full multi-head self-attention (MHSA) modules in DeiT-B~\cite{touvron2021deit} with our BRA and directly load the weights pre-trained on ImageNet before training on ADE20K dataset for semantic segmentation.
In this way, the linear projection weights of BRA modules are initialized with those of the original MHSA.
We compare such an adaptation with those proposed in~\cite{li2022vitdet}, \ie using local window attention (window size $w=14$) together with several global attention or convolution propagation blocks.
We set window size $w=4$ (which is equivalent to region partition size $S=8$ since the feature map has a resolution of $32 \times 32$) and the number of regions to attend $k=12$, hence each query attends to $4^2 \times 12 = 192$ key-value pairs, which is comparable to the local window attention where each query attends to $14 \times 14=196$ key-value pairs.

Table~\ref{table:adaptation} shows the results. Without propagation blocks, the architecture using BRA significantly surpasses the one with local window attention by 2.4 mAP. When further equipped with 4 global propagation blocks, the performance of both architectures is improved, while the one using BRA still has an advantage of 0.2 mAP.

\begin{table}[htpb]
\begin{center}
  \resizebox{0.95\linewidth}{!}{
  \begin{tabular}{l|c}
    \toprule
    attention function  &  mIoU(\%) \\
                        
    \midrule
     local window attention ($w=14$)  & 43.55 \\
     BRA($w=4, k=12$) & \textbf{45.92} \\
     \midrule
     local window attention + 4 conv prop. blks. & 44.68 \\
     local window attention + 4 global prop. blks. & 46.64 \\
     BRA + 4 global prop. blks. & \textbf{46.84} \\
    \bottomrule
    \end{tabular}
    }
\end{center}
\caption{Adapting pretrained ViT~\cite{vit} with BRA for semantic segmentation on ADE20K. For decoder, we use the Simple Feature Pyramid~\cite{li2022vitdet} followed by with Upernet~\cite{xiao2018upernet} head.}
\label{table:adaptation}
\end{table}

\section{More Visualization Results}

To further show how BRA works, we demonstrate more visualization results in Figure~\ref{fig:attn_vis_supp}.

{\small
\bibliographystyle{ieee_fullname}
\bibliography{egbib}

\begin{thebibliography}{10}\itemsep=-1pt

\bibitem{detr}
Nicolas Carion, Francisco Massa, Gabriel Synnaeve, Nicolas Usunier, Alexander
  Kirillov, and Sergey Zagoruyko.
\newblock End-to-end object detection with transformers.
\newblock In {\em European Conference on Computer Vision}, pages 213--229,
  2020.

\bibitem{chen2021regionvit}
Chun{-}Fu Chen, Rameswar Panda, and Quanfu Fan.
\newblock Regionvit: Regional-to-local attention for vision transformers.
\newblock In {\em The Tenth International Conference on Learning
  Representations, {ICLR} 2022, Virtual Event, April 25-29, 2022}.
  OpenReview.net, 2022.

\bibitem{chen2019mmdetection}
Kai Chen, Jiaqi Wang, Jiangmiao Pang, Yuhang Cao, Yu Xiong, Xiaoxiao Li,
  Shuyang Sun, Wansen Feng, Ziwei Liu, Jiarui Xu, et~al.
\newblock Mmdetection: Open mmlab detection toolbox and benchmark.
\newblock {\em arXiv:1906.07155}, 2019.

\bibitem{chen2022vitadapter}
Zhe Chen, Yuchen Duan, Wenhai Wang, Junjun He, Tong Lu, Jifeng Dai, and Yu
  Qiao.
\newblock Vision transformer adapter for dense predictions.
\newblock {\em arXiv preprint arXiv:2205.08534}, 2022.

\bibitem{chen2021dpt}
Zhiyang Chen, Yousong Zhu, Chaoyang Zhao, Guosheng Hu, Wei Zeng, Jinqiao Wang,
  and Ming Tang.
\newblock Dpt: Deformable patch-based transformer for visual recognition.
\newblock In {\em Proceedings of the 29th ACM International Conference on
  Multimedia}, pages 2899--2907, 2021.

\bibitem{child2019generating}
Rewon Child, Scott Gray, Alec Radford, and Ilya Sutskever.
\newblock Generating long sequences with sparse transformers.
\newblock {\em arXiv:1904.10509}, 2019.

\bibitem{chu2021twins}
Xiangxiang Chu, Zhi Tian, Yuqing Wang, Bo Zhang, Haibing Ren, Xiaolin Wei,
  Huaxia Xia, and Chunhua Shen.
\newblock Twins: Revisiting the design of spatial attention in vision
  transformers.
\newblock {\em Advances in Neural Information Processing Systems},
  34:9355--9366, 2021.

\bibitem{contributors2020mmseg}
MMSegmentation Contributors.
\newblock Openmmlab semantic segmentation toolbox and benchmark.
\newblock \url{https://github.com/open-mmlab/mmsegmentation}, 2020.

\bibitem{cubuk2020randaugment}
Ekin~D Cubuk, Barret Zoph, Jonathon Shlens, and Quoc~V Le.
\newblock Randaugment: Practical automated data augmentation with a reduced
  search space.
\newblock In {\em Proceedings of the IEEE/CVF Conference on Computer Vision and
  Pattern Recognition workshops}, pages 702--703, 2020.

\bibitem{dai2017deformable_conv}
Jifeng Dai, Haozhi Qi, Yuwen Xiong, Yi Li, Guodong Zhang, Han Hu, and Yichen
  Wei.
\newblock Deformable convolutional networks.
\newblock In {\em Proceedings of the IEEE International Conference on Computer
  Vision}, pages 764--773, 2017.

\bibitem{dai2019transformerxl}
Zihang Dai, Zhilin Yang, Yiming Yang, Jaime~G. Carbonell, Quoc~Viet Le, and
  Ruslan Salakhutdinov.
\newblock Transformer-xl: Attentive language models beyond a fixed-length
  context.
\newblock In Anna Korhonen, David~R. Traum, and Llu{\'{\i}}s M{\`{a}}rquez,
  editors, {\em Proceedings of the Conference of the Association for
  Computational Linguistics, {ACL} 2019, Volume 1: Long Papers}, pages
  2978--2988, 2019.

\bibitem{deng2009imagenet}
Jia Deng, Wei Dong, Richard Socher, Li-Jia Li, Kai Li, and Li Fei-Fei.
\newblock Imagenet: A large-scale hierarchical image database.
\newblock In {\em 2009 IEEE Conference on Computer Vision and Pattern
  Recognition}, pages 248--255, 2009.

\bibitem{devlin2018bert}
Jacob Devlin, Ming-Wei Chang, Kenton Lee, and Kristina Toutanova.
\newblock {BERT}: Pre-training of deep bidirectional transformers for language
  understanding.
\newblock In {\em Proceedings of the Conference of the North {A}merican Chapter
  of the Association for Computational Linguistics: Human Language
  Technologies, Volume 1 (Long and Short Papers)}, pages 4171--4186, June 2019.

\bibitem{cswin}
Xiaoyi Dong, Jianmin Bao, Dongdong Chen, Weiming Zhang, Nenghai Yu, Lu Yuan,
  Dong Chen, and Baining Guo.
\newblock Cswin transformer: A general vision transformer backbone with
  cross-shaped windows.
\newblock In {\em Proceedings of the IEEE/CVF Conference on Computer Vision and
  Pattern Recognition}, pages 12124--12134, 2022.

\bibitem{vit}
Alexey Dosovitskiy, Lucas Beyer, Alexander Kolesnikov, Dirk Weissenborn,
  Xiaohua Zhai, Thomas Unterthiner, Mostafa Dehghani, Matthias Minderer, Georg
  Heigold, Sylvain Gelly, Jakob Uszkoreit, and Neil Houlsby.
\newblock An image is worth 16x16 words: Transformers for image recognition at
  scale.
\newblock In {\em International Conference on Learning Representations, {ICLR}
  2021, 2021}, 2021.

\bibitem{goyal2017accurate_warmup}
Priya Goyal, Piotr Doll{\'a}r, Ross Girshick, Pieter Noordhuis, Lukasz
  Wesolowski, Aapo Kyrola, Andrew Tulloch, Yangqing Jia, and Kaiming He.
\newblock Accurate, large minibatch sgd: Training imagenet in 1 hour.
\newblock {\em arXiv:1706.02677}, 2017.

\bibitem{gupta2021topk}
Ankit Gupta, Guy Dar, Shaya Goodman, David Ciprut, and Jonathan Berant.
\newblock Memory-efficient transformers via top-k attention.
\newblock In Nafise~Sadat Moosavi, Iryna Gurevych, Angela Fan, Thomas Wolf,
  Yufang Hou, Ana Marasovic, and Sujith Ravi, editors, {\em Proceedings of the
  Workshop on Simple and Efficient Natural Language Processing, 2021}, pages
  39--52, 2021.

\bibitem{he2017maskrcnn}
Kaiming He, Georgia Gkioxari, Piotr Doll{\'a}r, and Ross Girshick.
\newblock Mask r-cnn.
\newblock In {\em Proceedings of the IEEE International Conference on Computer
  Vision}, pages 2961--2969, 2017.

\bibitem{he2016resnet}
Kaiming He, Xiangyu Zhang, Shaoqing Ren, and Jian Sun.
\newblock Deep residual learning for image recognition.
\newblock In {\em Proceedings of the IEEE Conference on Computer Vision and
  Pattern Recognition}, pages 770--778, 2016.

\bibitem{ho2019seqaxial}
Jonathan Ho, Nal Kalchbrenner, Dirk Weissenborn, and Tim Salimans.
\newblock Axial attention in multidimensional transformers.
\newblock {\em arXiv:1912.12180}, 2019.

\bibitem{huang2016stochastic_depth}
Gao Huang, Yu Sun, Zhuang Liu, Daniel Sedra, and Kilian Weinberger.
\newblock Deep networks with stochastic depth.
\newblock In {\em European Conference on Computer Vision}, pages 646--661,
  2016.

\bibitem{huang2019ccnet}
Zilong Huang, Xinggang Wang, Lichao Huang, Chang Huang, Yunchao Wei, and Wenyu
  Liu.
\newblock Ccnet: Criss-cross attention for semantic segmentation.
\newblock In {\em Proceedings of the IEEE/CVF International Conference on
  Computer Vision}, pages 603--612, 2019.

\bibitem{jiang2021token_labeling}
Zi-Hang Jiang, Qibin Hou, Li Yuan, Daquan Zhou, Yujun Shi, Xiaojie Jin, Anran
  Wang, and Jiashi Feng.
\newblock All tokens matter: Token labeling for training better vision
  transformers.
\newblock {\em Advances in Neural Information Processing Systems},
  34:18590--18602, 2021.

\bibitem{kirillov2019semanticfpn}
Alexander Kirillov, Ross Girshick, Kaiming He, and Piotr Doll{\'a}r.
\newblock Panoptic feature pyramid networks.
\newblock In {\em Proceedings of the IEEE/CVF Conference on Computer Vision and
  Pattern Recognition}, pages 6399--6408, 2019.

\bibitem{li2022uniformer}
Kunchang Li, Yali Wang, Peng Gao, Guanglu Song, Yu Liu, Hongsheng Li, and Yu
  Qiao.
\newblock Uniformer: Unified transformer for efficient spatiotemporal
  representation learning.
\newblock {\em arXiv:2201.04676}, 2022.

\bibitem{li2022vitdet}
Yanghao Li, Hanzi Mao, Ross Girshick, and Kaiming He.
\newblock Exploring plain vision transformer backbones for object detection.
\newblock {\em arXiv preprint arXiv:2203.16527}, 2022.

\bibitem{lin2017retinanet}
Tsung-Yi Lin, Priya Goyal, Ross Girshick, Kaiming He, and Piotr Doll{\'a}r.
\newblock Focal loss for dense object detection.
\newblock In {\em Proceedings of the IEEE International Conference on Computer
  Vision}, pages 2980--2988, 2017.

\bibitem{lin2014coco}
Tsung-Yi Lin, Michael Maire, Serge Belongie, James Hays, Pietro Perona, Deva
  Ramanan, Piotr Doll{\'a}r, and C~Lawrence Zitnick.
\newblock Microsoft coco: Common objects in context.
\newblock In {\em European conference on computer vision}, pages 740--755.
  Springer, 2014.

\bibitem{swin}
Ze Liu, Yutong Lin, Yue Cao, Han Hu, Yixuan Wei, Zheng Zhang, Stephen Lin, and
  Baining Guo.
\newblock Swin transformer: Hierarchical vision transformer using shifted
  windows.
\newblock In {\em Proceedings of the IEEE/CVF International Conference on
  Computer Vision}, pages 10012--10022, 2021.

\bibitem{loshchilov2017adamw}
Ilya Loshchilov and Frank Hutter.
\newblock Decoupled weight decay regularization.
\newblock In {\em 7th International Conference on Learning Representations,
  {ICLR} 2019, New Orleans, LA, USA, May 6-9, 2019}. OpenReview.net, 2019.

\bibitem{nvidia_blog}
Nvidia.
\newblock How to access global memory efficiently in cuda c/c++ kernels.
\newblock
  \url{https://developer.nvidia.com/blog/how-access-global-memory-efficiently-cuda-c-kernels/}.
\newblock Accessed: 2022-10-25.

\bibitem{paszke2019pytorch}
Adam Paszke, Sam Gross, Francisco Massa, Adam Lerer, James Bradbury, Gregory
  Chanan, Trevor Killeen, Zeming Lin, Natalia Gimelshein, Luca Antiga, et~al.
\newblock Pytorch: An imperative style, high-performance deep learning library.
\newblock In {\em Proceedings of Advances in Neural Information Processing
  Systems}, volume~32, 2019.

\bibitem{radford2018gpt}
Alec Radford, Karthik Narasimhan, Tim Salimans, and Ilya Sutskever.
\newblock Improving language understanding by generative pre-training.
\newblock 2018.

\bibitem{radosavovic2020regnet}
Ilija Radosavovic, Raj~Prateek Kosaraju, Ross Girshick, Kaiming He, and Piotr
  Doll{\'a}r.
\newblock Designing network design spaces.
\newblock In {\em Proceedings of the IEEE/CVF Conference on Computer Vision and
  Pattern Recognition}, pages 10428--10436, 2020.

\bibitem{ramachandran2019stand_slidingattn}
Prajit Ramachandran, Niki Parmar, Ashish Vaswani, Irwan Bello, Anselm Levskaya,
  and Jon Shlens.
\newblock Stand-alone self-attention in vision models.
\newblock In {\em Proceedings of Advances in Neural Information Processing
  Systems}, volume~32, 2019.

\bibitem{reddy2021dalle}
Mr~D~Murahari Reddy, Mr~Sk~Masthan Basha, Mr~M~Chinnaiahgari Hari, and Mr~N
  Penchalaiah.
\newblock Dall-e: Creating images from text.
\newblock 2021.

\bibitem{ren2022shunted}
Sucheng Ren, Daquan Zhou, Shengfeng He, Jiashi Feng, and Xinchao Wang.
\newblock Shunted self-attention via multi-scale token aggregation.
\newblock In {\em Proceedings of the IEEE/CVF Conference on Computer Vision and
  Pattern Recognition}, pages 10853--10862, 2022.

\bibitem{tang2022quadtree}
Shitao Tang, Jiahui Zhang, Siyu Zhu, and Ping Tan.
\newblock Quadtree attention for vision transformers.
\newblock In {\em The International Conference on Learning Representations,
  {ICLR} 2022, 2022}, 2022.

\bibitem{tay2020transformer_survey}
Yi Tay, Mostafa Dehghani, Dara Bahri, and Donald Metzler.
\newblock Efficient transformers: A survey.
\newblock {\em ACM Computing Surveys (CSUR)}, 2020.

\bibitem{touvron2021deit}
Hugo Touvron, Matthieu Cord, Matthijs Douze, Francisco Massa, Alexandre
  Sablayrolles, and Herv{\'e} J{\'e}gou.
\newblock Training data-efficient image transformers \& distillation through
  attention.
\newblock In {\em International Conference on Machine Learning}, pages
  10347--10357. PMLR, 2021.

\bibitem{tu2022maxvit}
Zhengzhong Tu, Hossein Talebi, Han Zhang, Feng Yang, Peyman Milanfar, Alan
  Bovik, and Yinxiao Li.
\newblock Maxvit: Multi-axis vision transformer.
\newblock In {\em ECCV}, 2022.

\bibitem{vaswani2017attention}
Ashish Vaswani, Noam Shazeer, Niki Parmar, Jakob Uszkoreit, Llion Jones,
  Aidan~N Gomez, {\L}ukasz Kaiser, and Illia Polosukhin.
\newblock Attention is all you need.
\newblock {\em Advances in neural information processing systems}, 30, 2017.

\bibitem{wang2020linformer}
Sinong Wang, Belinda~Z Li, Madian Khabsa, Han Fang, and Hao Ma.
\newblock Linformer: Self-attention with linear complexity.
\newblock {\em arXiv:2006.04768}, 2020.

\bibitem{wang2021pvt}
Wenhai Wang, Enze Xie, Xiang Li, Deng-Ping Fan, Kaitao Song, Ding Liang, Tong
  Lu, Ping Luo, and Ling Shao.
\newblock Pyramid vision transformer: A versatile backbone for dense prediction
  without convolutions.
\newblock In {\em Proceedings of the IEEE/CVF International Conference on
  Computer Vision}, pages 568--578, 2021.

\bibitem{wang2022pvtv2}
Wenhai Wang, Enze Xie, Xiang Li, Deng-Ping Fan, Kaitao Song, Ding Liang, Tong
  Lu, Ping Luo, and Ling Shao.
\newblock Pvt v2: Improved baselines with pyramid vision transformer.
\newblock {\em Computational Visual Media}, 8(3):415--424, 2022.

\bibitem{wang2021crossformer}
Wenxiao Wang, Lu Yao, Long Chen, Binbin Lin, Deng Cai, Xiaofei He, and Wei Liu.
\newblock Crossformer: A versatile vision transformer hinging on cross-scale
  attention.
\newblock In {\em International Conference on Learning Representations,
  {ICLR}}, 2022.

\bibitem{rw2019timm}
Ross Wightman.
\newblock Pytorch image models.
\newblock \url{https://github.com/rwightman/pytorch-image-models}, 2019.

\bibitem{xia2022dat}
Zhuofan Xia, Xuran Pan, Shiji Song, Li~Erran Li, and Gao Huang.
\newblock Vision transformer with deformable attention.
\newblock In {\em Proceedings of the IEEE/CVF Conference on Computer Vision and
  Pattern Recognition}, pages 4794--4803, 2022.

\bibitem{xiao2018upernet}
Tete Xiao, Yingcheng Liu, Bolei Zhou, Yuning Jiang, and Jian Sun.
\newblock Unified perceptual parsing for scene understanding.
\newblock In {\em Proceedings of the European conference on computer vision
  (ECCV)}, pages 418--434, 2018.

\bibitem{ScalableViT}
Rui Yang, Hailong Ma, Jie Wu, Yansong Tang, Xuefeng Xiao, Min Zheng, and Xiu
  Li.
\newblock Scalablevit: Rethinking the context-oriented generalization of vision
  transformer.
\newblock {\em arXiv:2203.10790}, 2022.

\bibitem{yao2022wavevit}
Ting Yao, Yingwei Pan, Yehao Li, Chong-Wah Ngo, and Tao Mei.
\newblock Wave-vit: Unifying wavelet and transformers for visual representation
  learning.
\newblock In {\em European Conference on Computer Vision}, pages 328--345.
  Springer, 2022.

\bibitem{yun2019cutmix}
Sangdoo Yun, Dongyoon Han, Seong~Joon Oh, Sanghyuk Chun, Junsuk Choe, and
  Youngjoon Yoo.
\newblock Cutmix: Regularization strategy to train strong classifiers with
  localizable features.
\newblock In {\em Proceedings of the IEEE/CVF International Conference on
  Computer Vision}, pages 6023--6032, 2019.

\bibitem{zeng2022tcformer}
Wang Zeng, Sheng Jin, Wentao Liu, Chen Qian, Ping Luo, Wanli Ouyang, and
  Xiaogang Wang.
\newblock Not all tokens are equal: Human-centric visual analysis via token
  clustering transformer.
\newblock In {\em Proceedings of the IEEE/CVF Conference on Computer Vision and
  Pattern Recognition}, pages 11101--11111, 2022.

\bibitem{zhang2017mixup}
Hongyi Zhang, Moustapha Ciss{\'{e}}, Yann~N. Dauphin, and David Lopez{-}Paz.
\newblock mixup: Beyond empirical risk minimization.
\newblock In {\em International Conference on Learning Representations, {ICLR}
  2018}, 2018.

\bibitem{zhou2019ade20k}
Bolei Zhou, Hang Zhao, Xavier Puig, Tete Xiao, Sanja Fidler, Adela Barriuso,
  and Antonio Torralba.
\newblock Semantic understanding of scenes through the ade20k dataset.
\newblock {\em International Journal of Computer Vision}, 127(3):302--321,
  2019.

\end{thebibliography}
}

\end{document}